\documentclass[times, review, 10pt]{elsarticle}
\usepackage{caption}
\usepackage{subcaption}
\usepackage{bbding}
\usepackage{utfsym}
\usepackage{multirow}
\usepackage{graphicx}
\usepackage[graphicx]{realboxes}
\usepackage{amsmath}
\usepackage{amsfonts}
\usepackage{algorithm}
\usepackage[colorlinks,bookmarksopen,bookmarksnumbered,citecolor=red, linkcolor=red, urlcolor=red]{hyperref}

\usepackage{enumitem}
\usepackage{newfloat}
\usepackage{listings}
\usepackage{algpseudocode}
\journal{Elsevier }

\begin{document}

\begin{frontmatter}

\title{DFDNet: Dynamic Frequency-Guided De-Flare Network}

\author[1]{Minglong Xue} %% Author name
\author[1]{Aoxiang Ning}
\author[2]{Shivakumara Palaiahnakote}
\author[3]{Mingliang Zhou}
\affiliation[1]{organization={College of Computer Science and Engineering},%Department and Organization
            addressline={Chongqing University of Technology}, 
            city={Chongqing},
            postcode={400054}, 
            country={China}}

\affiliation[2]{organization={School of Science, Engineering and Environment},%Department and Organization
            addressline={ University of Salford}, 
            city={Manchester},
            % postcode={400044}, 
            country={UK}}
\affiliation[3]{organization={College of Computer Science},%Department and Organization
            addressline={Chongqing University}, 
            city={Chongqing},
            postcode={400044}, 
            country={China}}

\begin{abstract}
Strong light sources in nighttime photography frequently produce flares in images, significantly degrading visual quality and impacting the performance of downstream tasks. While some progress has been made, existing methods continue to struggle with removing large-scale flare artifacts and repairing structural damage in regions near the light source.
We observe that these challenging flare artifacts exhibit more significant discrepancies from the reference images in the frequency domain compared to the spatial domain. Therefore, this paper presents a novel dynamic frequency-guided deflare network (DFDNet) that decouples content information from flare artifacts in the frequency domain, effectively removing large-scale flare artifacts. Specifically, DFDNet consists mainly of a global dynamic frequency-domain guidance (GDFG) module and a local detail guidance module (LDGM). The GDFG module guides the network to perceive the frequency characteristics of flare artifacts by dynamically optimizing global frequency domain features, effectively separating flare information from content information. Additionally, we design an LDGM via a contrastive learning strategy that aligns the local features of the light source with the reference image, reduces local detail damage from flare removal, and improves fine-grained image restoration. The experimental results demonstrate that the proposed method outperforms existing state-of-the-art methods in terms of performance. The code is available at \href{https://github.com/AXNing/DFDNet}{https://github.com/AXNing/DFDNet}.
\end{abstract}

\end{frontmatter}



\section{Introduction}
\label{sec1}
%% Labels are used to cross-reference an item using \ref command.

% \begin{figure*}[ht!]
% \centering
%   \includegraphics[height=0.5\linewidth,width=\linewidth]{cyfig1.pdf}
%   \caption{CycleRDM is capable of generating high-fidelity restoration in a variety of tasks. CycleRDM gives faithful results on a wide range of {\bf (a)} linear image restoration tasks. Meanwhile, CycleRDM also realizes {\bf (b)} blind, non-linear image enhancement tasks with high quality.}
%   \label{fig:1}
% \end{figure*}
Lens flare is an optical artifact that occurs when a strong light source enters the camera lens. Scattering or reflection of light within the optical system leads to the formation of bright radial patterns, light spots, blurred areas, or streaks in the image. This effect is commonly observed in photography and computer vision, especially in nighttime settings, where multiple artificial light sources amplify the impact of flares. Lens flares not only reduce image contrast but also suppress details around the light source, leading to significant degradation in visual quality and affecting downstream computer vision tasks, such as face detection\cite{hai2023r2rnet, jin2021pedestrian}, semantic segmentation, and optical flow estimation\cite{zheng2020optical}.

% \begin{figure}[t]
%         \centering
%         \includegraphics[height=0.25\textwidth, width=0.4\textwidth]{tu1.pdf}
%         \caption{Comparison of our method with current state-of-the-art methods in terms of the G-PSNR and S-PSNR metrics on the FLAIR7k++ synthetic dataset.}
%         \label{tu1}
%     \vspace{-2em}
% \end{figure}
\begin{figure*}[t]
        \centering
        \includegraphics[height=0.4\textwidth, width=0.8\textwidth]{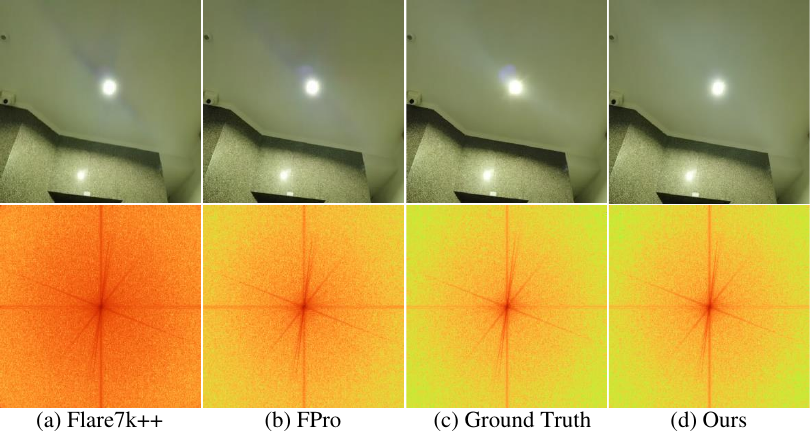}
        \caption{Comparison of restoration results on the Flare7k++ dataset using different methods: (a) results of Flare7k++~\cite{10541091}, (b) results of FPro\cite{zhou2024seeing}, (c) the ground truth, and (d) our method. The first row presents the reconstruction results in the spatial domain, whereas the second row displays the corresponding frequency spectra.
}
        \label{tu1}
\end{figure*}
% From the results, it can be observed that flare artifacts exhibit more significant differences in the frequency domain than in the spatial domain, deviating more noticeably from the reference image.

% Lens flare can be broadly classified into scattering flare and reflection flare. Scattering flares arise from dust, scratches, or wear on the lens surface. It typically manifests as radial streaks surrounding the light source and remains stationary relative to the light source, regardless of camera or light source movement. Reflection flare results from multiple internal reflections within the lens system. When the camera is adjusted, it appears as circular or polygonal patterns shaped by the light source and shifts opposite to the light source's movement.
Although lens antireflective coatings can somewhat mitigate this issue, contaminated lens surfaces in simplified systems, such as those found in smartphones, often worsen the problem. The development of a universal method for flare removal is particularly challenging because of the diversity in flare shapes, colours, and sizes. Moreover, their complex interactions with scene content make it difficult to differentiate genuine scene elements from flare-induced artifacts. Effortlessly removing flares while preserving overall image quality remains an open challenge in this field.\par

Traditional flare removal methods \cite{chabert2015automated, asha2019auto} are typically divided into two stages: detection and removal. These methods first estimate the shape and location of the flare and then use sample patches to restore the affected regions. However, flares in real-world scenarios exhibit diverse shapes and patterns, making traditional methods ineffective at handling complex cases. Recently, several deep learning-based methods\cite{guo2025underwater, yu2024multi} have been proposed. Wu et al.\cite{wu2021train} synthesized flare-damaged images by directly incorporating flare patterns into scene images to train neural networks. To address the model's poor performance under nighttime conditions, Flare7k \cite {dai2022flare7k} introduced a new dataset for removing nighttime flares. While these methods improve the model's flare removal capability from a data-driven perspective, they still face challenges such as image blurriness and ghosting artifacts.
To further enhance the quality of the restored image, Harmonizing \cite{qu2024harmonizing} introduced a plug-and-play adaptive focusing module that adaptively masks clean background areas, allowing the model to focus on regions severely affected by flares. However, these methods fail to effectively separate content information from artifact regions, limiting their restoration capacity in large-scale flare areas.\par

% Flare artifacts can occupy a large portion of an image or even the entire image, and their frequency characteristics often differ from those of normal images. Therefore, from the perspective of the global frequency domain, we design a dynamic frequency-domain guidance module that dynamically optimizes the global frequency domain weights of the features to separate flare artifacts from content information, guiding the network to preserve content information while removing the artifacts.
\begin{figure*}[t]
        \centering
        \includegraphics[height=0.26\textwidth, width=0.98\textwidth]{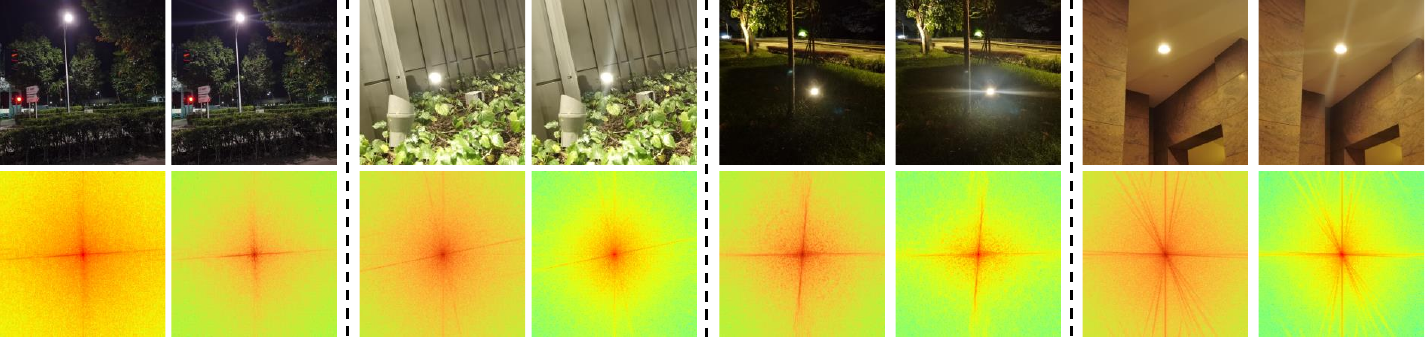}
        \caption{Comparison of flare-corrupted images and their corresponding reference images in the frequency domain. For each pair of images, the image on the left is the reference image, and the image on the right is the image corrupted by flare. The interference of flares on images shows significant differences in the frequency domain.}
        \label{tu2} 
    \end{figure*}
We transform the images restored by existing methods and the reference images from the spatial domain to the frequency domain, as shown in Fig. \ref{tu1}. Although existing methods effectively remove prominent flares in the spatial domain, they still struggle to eliminate flare artifacts, which become even more pronounced in the frequency domain. Furthermore, as illustrated in Fig. \ref{tu2}, we compare the frequency domain representations of various flare-damaged images with the reference images. The spectrum of the GT image is concentrated in the low-frequency region, appearing darker overall, with dispersed and naturally distributed dot-like patterns in the peripheral areas. This reflects the frequency characteristics of details and textures in natural images. In contrast, the spectrum of the flare-affected image shows a significant increase in overall brightness, especially with distinct radial stripe patterns along specific directions. This finding indicates that the intense spatial domain fluctuations caused by flares introduce substantial interference to the mid- and high-frequency components. These pronounced differences in the frequency domain facilitate the detection and differentiation of flare artifacts, providing more discriminative features for their removal. Based on these findings, we propose the dynamic frequency-guided deflare network (DFDNet), with the global dynamic frequency-domain guidance (GDFG) module as its core component. To address the limited global perception capability of the window attention mechanism in the existing U-former architecture, the GDFG module introduces a dual-domain analysis framework that operates in both the spatial and frequency domains by utilizing the discrete Fourier transform. It incorporates a learnable frequency domain weight modulation mechanism that dynamically generates frequency weights from multichannel features, enabling precise decoupling and suppression of flare artifact characteristics. This design allows the model to adaptively identify abnormal frequency components while preserving the inherent frequency features of background textures, thereby enhancing the deflare performance at both the global and local levels. Additionally, we propose a contrastive learning strategy to align local regions around the light source with reference images, effectively mitigating local detail degradation induced by flare removal. \par
Compared to existing methods, our DFDNet achieves superior flare removal by jointly leveraging both global and local perspectives. Globally, the GDFG module operates in the frequency domain to capture long-range contextual dependencies and suppress widespread flare artifacts, effectively addressing the limitations of local window-based attention. Locally, the contrastive learning strategy guide the network to focus on subtle texture differences and structural details in the vicinity of light sources. The integration of these two complementary perspectives enables our model to isolate and eliminate flare artifacts more accurately, while preserving fine-grained scene textures.\par
In summary, the contributions of this paper can be summarized as follows:
\begin{itemize}
\item We propose a novel dynamic frequency-guided deflare network that effectively removes flares at both the global and local levels.
\item We present a global dynamic frequency-domain guidance module that decouples flare artifacts from content information in the frequency domain.
\item We design a local detail guidance module to align local regions near the light source with reference images, effectively mitigating local detail degradation caused by flare removal.
\item Extensive experiments conducted on benchmark datasets demonstrate the effectiveness of our proposed method.
\end{itemize}

The remainder of this paper is structured as follows. In Section \ref{Related Works}, related works are discussed. In Section \ref{Method}, the proposed novel model method is described in detail. The relevant experimental setup and results are shown in Section \ref{EXPERIMENTS}. In Section \ref{Limitations}, we discuss the method's limitations and future work. Section \ref{Conclusions} presents the conclusions.

\section{Related work}\label{Related Works}

\subsection{Flare Remove}
\subsubsection{Physics-Based Flare Removal Methods}
The most common optical solution to avoid lens flares is to coat the lens surface with an antireflective coating\cite{chen2010antireflection}. This method leverages the principle of destructive interference to reduce reflections within the lens system, significantly enhancing light transmission. However, it cannot completely eliminate reflections and tends to fail when the light source is extremely bright. Another common approach is to improve the material of camera lenses to reduce flare artifacts. For example, Boynton et al. \cite{boynton2003liquid} proposed a liquid-filled camera lens to mitigate flare artifacts caused by light reflections. McLeod et al. \cite{macleod2010thin} utilized a neutral density filter to minimize reflective flare. Although these specific physical methods can eliminate certain lens flare artifacts, they generally fail to address unforeseen flare issues in complex environments.

\subsubsection{Deep Learning-based Flare Removal Methods}
Early methods\cite{chabert2015automated, asha2019auto, vitoria2019automatic} for flare removal were mostly two-stage approaches: detection and repair. These methods rely on strong assumptions about the illumination, shape, and position of the flare for detection, followed by patch-based techniques to repair the affected regions. For example, Chabert et al.\cite{chabert2015automated} used a series of thresholds to binarize an image, calculated the contour features of the binarized image, and identified potential flare candidate regions for reconstruction. Similarly, Vittoria et al.\cite{vitoria2019automatic} detected flares by analysing overexposed regions near flare points and creating flare masks to remove them. However, these handcrafted feature-based methods are only effective for limited types of flares, often misidentifying locally bright areas as flares and struggling to differentiate between various flare types.\par
Recently, data-driven learning approaches\cite{feng2021removing, feng2023generating, zhang2023ff, zhou2023improving, song2023hard, niu2024gr} have been proposed. Wu et al.\cite{wu2021train} introduced a synthesis method for paired training data, where flare images were directly added to scene images to simulate flare-corrupted images for training neural networks. However, their method did not generalize well to real-world data. Qiao et al.\cite{qiao2021light} proposed a network trained on unpaired flare data consisting of a light source detection module, a flare detection and removal module, and a generation module. Dai et al. \cite{dai2022flare7k} created the first benchmark dataset for night flare removal, the Flare7K dataset, which provides a valuable baseline for studying this challenging task. Since artificial and solar light spectra produce different diffraction patterns, the Flare7K++\cite{10541091} dataset enhances the synthetic Flare7K\cite{dai2022flare7k} dataset with new real-world captured data from the Flare-R dataset. To further improve image restoration quality, MFDNet\cite{jiang2024mfdnet} proposed a lightweight multifrequency delayed network based on a Laplacian pyramid, decomposing flare-corrupted images into low- and high-frequency bands to separate lighting from content information effectively.
% Although these methods can achieve preliminary flare removal, they fail to effectively separate content information from artifact regions, limiting their ability to restore large-scale artifact areas. Additionally, these approaches often damage the content near light sources, resulting in degraded local details in such regions.

\subsection{Applications of the Frequent Analysis in Low-Level Vision}
Frequency domain analysis has been widely applied in various low-level vision tasks\cite{sun2025self, song2025efficient}, such as image restoration \cite{wu2025freprompter, gao2025frequency} and low-light image enhancement \cite{ning2025kan, he2024optimizing, xu2022illumination}, owing to its ability to effectively separate and process different frequency components of images. By transforming images from the spatial domain to the frequency domain, it becomes easier and more efficient to manipulate fine-grained details, suppress noise, and enhance structural information. This property is particularly advantageous for tasks that require precise texture restoration or illumination correction.\par
For example,
% DFFormer \cite{tatsunami2024fft} leverages the Fourier transform to design a global filtering mechanism, enabling the model to capture long-range dependencies with lower computational cost than conventional convolutional approaches while maintaining excellent performance in image restoration tasks. Similarly,
Huang et al. \cite{huang2024wavedm} proposed a wavelet-based diffusion model (WaveDM), which learns the distribution of clean images in the frequency domain after performing wavelet decomposition. This design not only improves inference efficiency by simplifying the data distribution but also enhances the model's ability to reconstruct multiscale frequency components, leading to superior restoration quality.\par
Additionally, Cui et al. \cite{cui2023image} introduced a simple yet effective discriminative frequency-domain network based on the fast Fourier transform (FFT). Their method exploits frequency-aware features to improve image restoration performance while maintaining computational efficiency. Qiao et al.\cite{qiao2023learning} designed a spatial‒frequency interaction residual block (SFIR), which efficiently learns global frequency information and local spatial features in an interactive manner. MFSNet\cite{10747495} is a multiscale frequency selection framework that integrates spatial and frequency domain features for image restoration. It employs dynamic filter selection (DFS) modules and frequency cross-attention mechanisms (FCAMs) to adaptively extract high- and low-frequency information. Although these works demonstrate notable advantages in preserving image structures, improving restoration accuracy, and accelerating inference, they generally rely on fixed or manually designed filters and lack learnable, content-adaptive dynamic filtering mechanisms. In contrast, our method introduces a learnable multi-channel frequency modulation strategy, which adaptively adjusts frequency responses based on input characteristics.

\begin{figure*}[t]
        \centering
        \includegraphics[height=0.5\textwidth, width=0.98\textwidth]{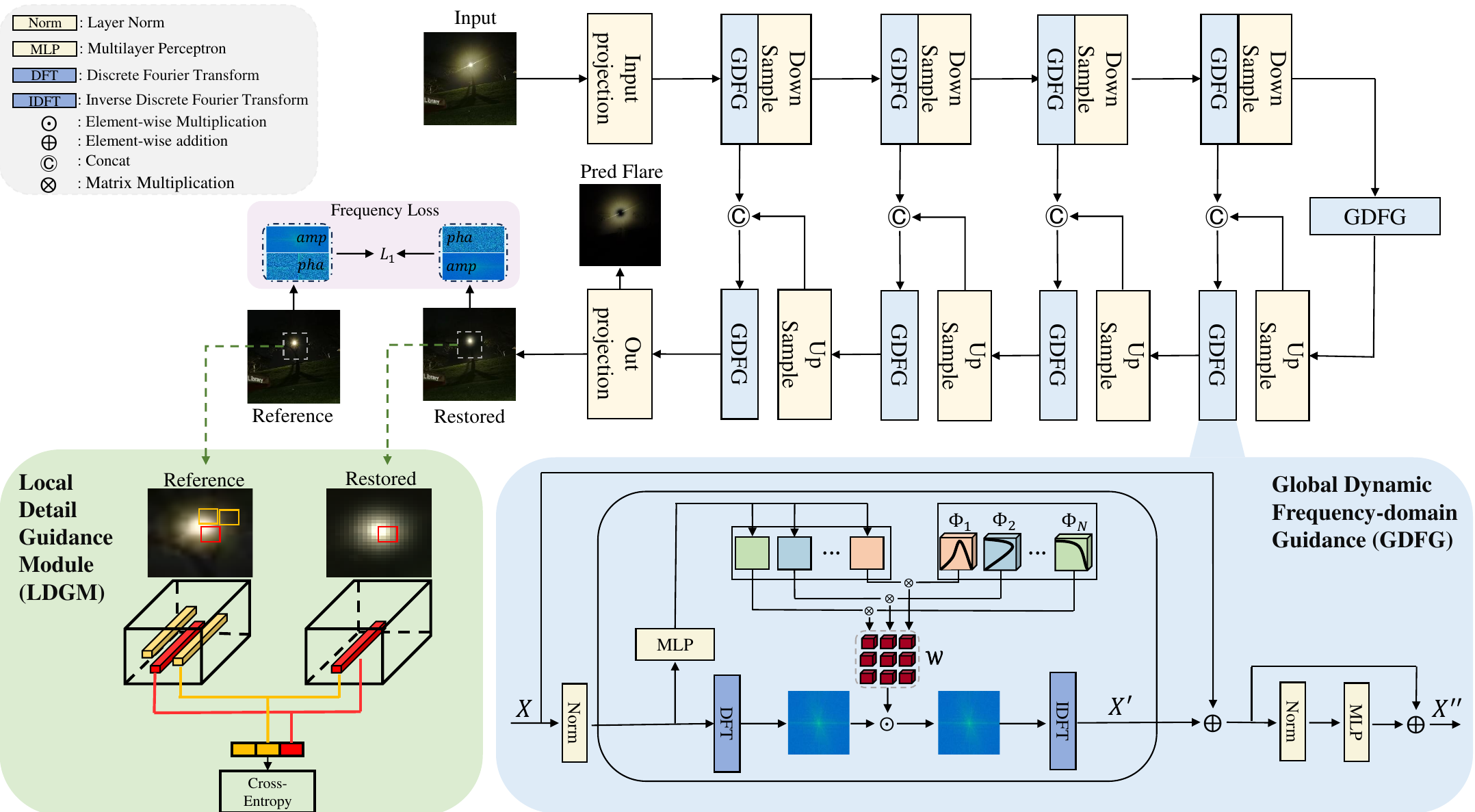}
        \caption{Overview of DFDNet. DFDNet consists of multiple global dynamic frequency-domain guidance (GDFG) modules.}
        \label{framework}
    % \vspace{-2em}
    \end{figure*}
\section{Method}\label{Method}
 In this section, we first describe the overall pipeline and architecture of the proposed flare removal network, DFDNet. Then, we provide detailed information about the core component of DFDNet, the global dynamic frequency-domain guidance module. Finally, we introduce the local detail guidance module.
 
\subsection{Overall Pipeline}
The Uformer-based method\cite{wang2022uformer} has achieved some progress in flare removal, which progressively extracts global features in the feature space through multiple downsamplings. However, because of the local window attention mechanism, it inevitably overlooks some global information and struggles to analyse the characteristics of flares and normal texture images effectively in the spatial domain. Therefore, we propose a dynamic frequency-guided deflare network (DFDNet) to explore flare removal from both global and local perspectives. \par
Specifically, as shown in Fig. \ref{framework}, given a flare-damaged image $I$, we first apply a 3 $\times$ 3 convolution layer with LeakyReLU activation to extract low-level features $E_0\in \mathbb{R}^{C \times H \times W}$. Next, these features pass through four encoders. Each encoder substage consists of a GDFG module and a downsampling layer. The GDFG module transforms the features $E_{i-1}$ into the frequency domain and dynamically learns the frequency characteristics of the flare artifacts. In the downsampling layer, we first reshape the flattened features into 2D spatial feature maps and then downsample the maps by applying a 4 $\times$ 4 convolution. Each encoder stage can be represented as:
\begin{equation}
\begin{aligned}
E_{i}&= Encoder(E_{i-1}) \\
 &= Down(GDFG(E_{i-1})),
 \end{aligned}
\end{equation}
where $i\in 1 \sim 4$; $GDFG$ is the global dynamic frequency-domain guidance module; and $Down$ is the downsampling layer. We employ the GDFG module as the bottleneck stage of the network to isolate flare features within the latent space. During the decoding stage, the image is reconstructed via a combination of the GDFG module and upsampling layers. Upsampling is initially performed via a 2 $\times$ 2 transposed convolution with a stride of 2, which halves the number of feature channels while doubling the spatial resolution of the feature maps. The upsampled feature maps $D_{i}$ are then skip-connected with the corresponding encoder features $E_{i}$ and passed into the GDFG module for further processing. The process can be expressed as:
\begin{equation}
\begin{aligned}
D_{i-1}&= Decoder(D_{i}) \\
 &= GDFG(Cat(Up(D_{i}), E_{i})),
 \end{aligned}
\end{equation}
where $Up$ is the upsampling layer and where $Cat$ denotes Conact. Following the decoder stages, the flattened features are reshaped into 2D feature maps, and a 3 $\times$ 3 convolutional layer is applied to produce the restored image and the predicted flare map. To mitigate the impact of localized regions near the light source during flare removal, we propose a contrastive learning strategy and design a local detail guidance module (LDGM). This approach facilitates positively oriented refinement by maximizing the mutual information between the light source regions in the recovered image and the reference image.
\begin{figure*}[t]
        \centering
        \includegraphics[height=0.18\textwidth, width=\textwidth]{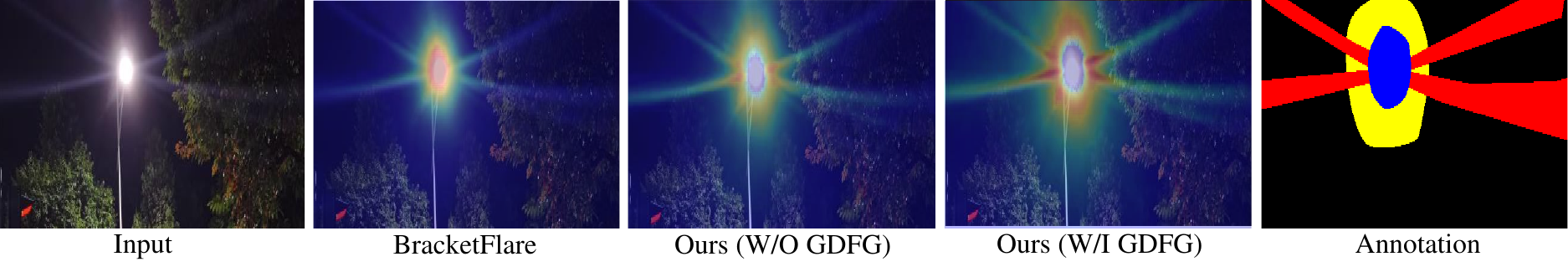}
        \caption{Comparison of the predicted flare results with and without the GDFG module. The last column represents the annotation information, where glare, streak, and light sources are marked in yellow, red, and blue, respectively.}
        \label{filter}
    \end{figure*}
\subsection{Global Dynamic Frequency-domain Guidance Module}
Early global filter methods use learnable global filter weights to optimize and adjust the Fourier spectrum.
% which is defined as:
% \begin{equation}
% \mathbb{G}=\mathcal{F}^{-1}(K\odot \mathcal{F}(X))
% \end{equation}
% where $\mathcal{F}$ represents the 2D discrete Fourier transform, $\mathcal{F}^{-1}$ denotes the 2D discrete inverse Fourier transform, and $K$ is a learnable parameter.
Since large flare artifact features are deeply fused with background information, separating these features via a single global filter is challenging. Therefore, we formulate N-dimensional global dynamic weights, which are applied to each channel of the frequency domain features. The coefficients of these weights are determined by the input features, enabling the model to perceive the target features more quickly. \par
Specifically, we first define the multichannel global filter:
\begin{equation}
\mathcal{G} =\mathcal{F}^{-1}(\mathcal{W} \odot \mathcal{F}(X)),
\end{equation}
where $\mathcal{W}$ represents multichannel dynamic weights. This procedure is shown in Algorithm \ref{processcode}. First, we define $N$ learnable parameters $\Phi =\left \{ \Phi _1,..., \Phi_N\right \} $ as the initial filters. Next, we compute the dynamic weight coefficients of the filters. Specifically, we perform global average pooling on the spatial dimensions of the input feature map $X\in \mathbb{R}^{C \times H \times W}$ to obtain the global features $\widehat{X}\in \mathbb{R}^C$ of each channel, where $\hat{X}_c=\frac{1}{HW} {\textstyle \sum_{h=1}^{H}} {\textstyle \sum_{w=1}^{W}}X_{c,h,w}$ and $c \in \left \{ 1,..., C \right \} $ is the channel index. These channel features are then processed through a multilayer perceptron (MLP) to generate weighted representations $s\in \mathbb{R}^{N \times e^{'} }$ for each channel, where $e^{'}$ is the hidden layer dimension of the MLP. The MLP is defined as:
\begin{equation}
\mathcal{M}(\mathbf{X}) = W_2 \cdot StarReLU(W_1 \cdot LN(\mathbf{X})),
\end{equation}
where $W_1 \in \mathbb{R}^{e^{'} \times C}$ and $W_2 \in \mathbb{R}^{N\times e^{'}}$ are learnable weight matrices, $LN$ denotes layer normalization, and $StarReLU$ is the activation function. We then reshape $s$ into $N$ weight coefficients $\mathcal{T}_{i}(X_{c,:,:})=  \frac{e^{s(c-1)N+i}}{\sum_{n=1}^{N} e^{s(c-1)N+n}} $, where $i \in \left \{ 1,..., N \right \} $, and normalize them along the channel dimension via the Softmax operation to generate dynamic weight coefficients. These coefficients are elementwise multiplied with the initial filter weights $\Phi _i$ to obtain the dynamic weights $\mathcal{W} =\sum_{i=0}^{N} \mathcal{T}_{i}(X_{c,:,:})\Phi _{i}$. This process is implemented in lines 10--14 of Algorithm \ref{processcode}.
\begin{algorithm}[t]
\caption{Dynamic Optimization Weights Mechanism}
\label{processcode}
\begin{algorithmic}[1]
\State \textbf{Input:} Feature map $X \in \mathbb{R}^{C \times H \times W}$
\State \textbf{Output:} Output feature map $X'$
\State \textbf{DFT:}
\Statex \hspace{1em} $\tilde{x}(h^{'},w^{'}) = \sum_{h=0}^{H-1} \sum_{w=0}^{W-1} \frac{x(h,w)e^{-2\pi j\left(\frac{h h^{'}}{H}+\frac{w w^{'}}{W} \right)}}{\sqrt{HW}}$
\State \textbf{IDFT:}
\Statex \hspace{1em} $x(h,w) = \sum_{h'=0}^{H-1} \sum_{w'=0}^{W-1} \tilde{x}(h',w') e^{2\pi j \left(\frac{h h'}{H} + \frac{w w'}{W} \right)} \frac{1}{\sqrt{HW}}$
\State \textbf{Process:}
\Statex \hspace{1em} $\mathcal{W} = \sum_{i=1}^{N} \mathcal{T}_i(X_{c,:,:}) \Phi_i$
\Statex \hspace{1em} $(\Phi_1, \ldots, \Phi_N) \in \mathbb{C}^{H \times \left\lceil \frac{W}{2} \right\rceil}$
\Statex \hspace{1em} $\mathcal{T}_i(X_{c,:,:}) = \frac{e^{s(c-1)N + i}}{\sum_{n=1}^{N} e^{s(c-1)N + n}}$
\Statex \hspace{1em} $(s_1, \ldots, s_{Nc'})^\top = \mathcal{M} \left( \frac{1}{HW} \sum_{h,w} \mathbf{X}_{:,h,w} \right)$
\State \textbf{Output:}
\Statex \hspace{1em} $X' = \text{IDFT}(\mathcal{W} \odot \text{DFT}(X))$
\end{algorithmic}
\end{algorithm}
% Subsequently, we apply the Softmax operation along the channel dimension to convert the channel information into weight coefficients, which are then multiplied elementwise with the initial weights of the filters. This process is implemented in lines 10--15 of Algorithm \ref{processcode}.\par
After obtaining the dynamic weights, we apply the discrete Fourier transform to the input feature map $X$, transform the features from the spatial domain to the frequency domain, and then apply the obtained weights to the frequency-domain features. Next, we perform the inverse discrete Fourier transform to convert the features back to the spatial domain. The mathematical expression for this process is on line 16 of Algorithm \ref{processcode}. Finally, we perform a residual connection with the input feature map to preserve important details. The whole process can be represented as:
\begin{equation}
\begin{aligned}
X^{''} &= GDFG(X)\\
       &=LN(MLP(\mathcal{G}(X) +X))+ \mathcal{G}(X) +X.
\end{aligned}
\end{equation}
Fig. \ref{filter} compares the predicted flare results with and without the GDFG module. The results clearly demonstrate that the module can effectively extract streak and glare regions.

\subsection{ Local detail guidance module}
\begin{figure}[t]
        \centering
        \includegraphics[height=0.5\textwidth, width=0.6\textwidth]{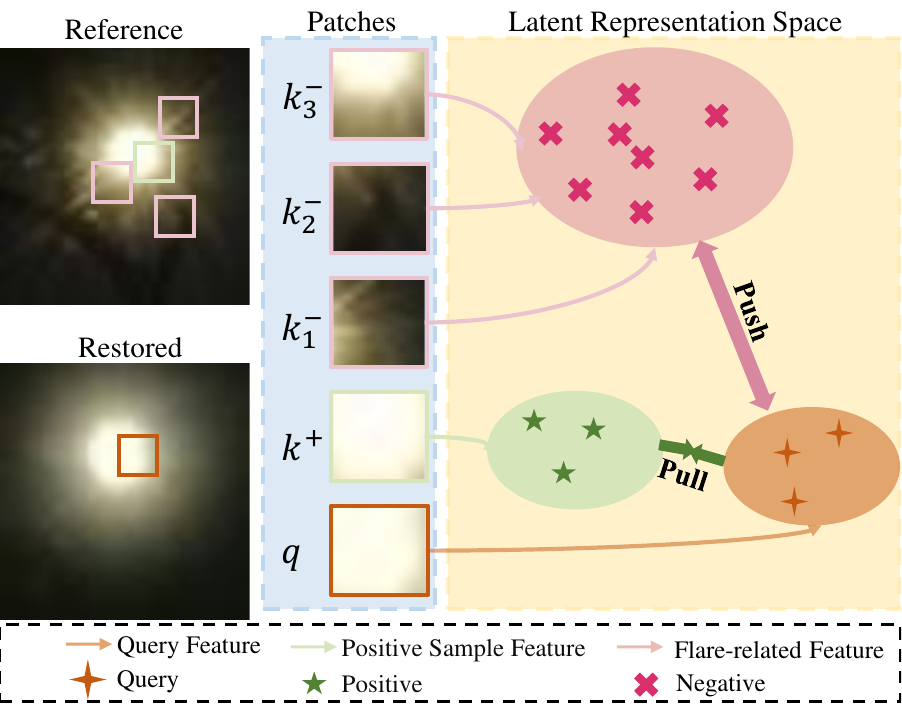}
        \caption{Illustration of the LDGM, where $q$, $k^+$, and $k^{-}_{n}$ represent the query samples, positive samples, and negative samples, respectively.}
        \label{cut}
    \end{figure}

The previous method leads to a loss of localized details of the light source when the flare is removed. To address this issue, we design a local detail guidance module (LDGM) based on a contrastive learning framework. As illustrated in Fig. \ref{cut}, we adopt a patch-based approach instead of processing the entire image. This method focuses on aligning local details by constraining the local consistency between the features of the target and reference images, which also accelerates model convergence.\par
Specifically, given the output image $\hat{I}_{0}$ from the restoration network and the reference image $I_{0}$, we first convert them into the two-dimensional spatial domain and randomly extract local feature patches. These smaller feature subsets are then used for contrastive learning. Contrastive learning involves three signals: a query sample $q$, a positive sample $k^{+}$ and a negative sample $k^{-}$. Our objective is to clearly distinguish the central region of the light source from other patches in the input, such as halos around the light source or dark background areas, ensuring tighter alignment between these regions and the light source during local feature matching. Therefore, as illustrated in Fig. \ref{cut}, we use patches from the restored image $\hat{I}_{0}$ as the query sample $q$ and calculate the corresponding patches from the reference image $I_{0}$ as the positive sample $k^{+}$. The similarity is measured as follows:
\begin{equation}
sim(z_1,z_2)=\frac{z_{1}\cdot z_{2}}{\left \| z_1 \right \|\left \| z_2 \right \|  } .
\end{equation}
The remaining patches are used as the negative sample $k^-$. Next, we map $q$, $k^{+}$ and $k^{-}_{N}$ into $K$-dimensional vectors $v$, $v^{+}\in \mathbb{R}^{K} $, and $v^{-}\in \mathbb{R}^{N\times K}$, respectively. $v_{n}^{-}$ denotes the $n$-th negative sample. We normalize these samples onto a unit sphere to prevent spatial expansion or collapse. Finally, we define the loss function $L_{LDG}$, which maximizes the similarity between positive samples and minimizes the similarity between negative samples, ensuring that the local light source of the restored image aligns with the reference image. The mathematical expression for $L_{LDG}$ is as follows:
% The center region of the light source should be clearly distinguished from other patches in the input (e.g., the halo around the light source or the dark background), as shown in Fig. \ref{cut}, ensuring that these regions are more closely aligned with the light source when local features are matched. Therefore, we construct a set of negative samples from other feature blocks in the batch and define the process as follows:
% \begin{equation}
%  L_{patch} = \sum_{s=1}^{S} l(v,v^{+},v^{-}_{s}) ,
% \end{equation}
% where $v$ represents a patch from the restored image, $v^{+}$ denotes the corresponding patch in the reference image, and $v^{-}$ represents the negative patch.
%(v,v^{+},v^{-}_{s})
\begin{equation}
L_{LDG} =-log\left [ \frac{exp(v\cdot v^{+}/\tau)}{exp(v\cdot v^{+}/\tau +\sum_{n=1}^{N}exp(v\cdot v_{n}^{-}/\tau)  )}  \right ],
\end{equation}
where $\tau$ is a temperature hyperparameter that controls the scaling factor of similarity scores in contrastive loss. It influences the sharpness of the Softmax distribution and adjusts the discriminative margin between positive and negative sample similarities.

\subsection{Loss Function}
During training, we first add the flare image $F$ to the background image $I_{0}$ to obtain the flare-damaged image $I$. Our flare removal network, denoted as $\Psi $, takes the flare-damaged image $I$ as input. The estimated flare-free image $I_0$ and the flare image $F$ can then be expressed as:
\begin{equation}
\hat{I}_{0},\hat{F}  =\Psi  (I).
\end{equation}
We combine $L_1$ loss and $MSE$ loss as the perceptual loss to supervise the flare and background images, minimizing the content difference between the restored image $\hat{I}_{0}$ and the reference image $I_{0}$. The perceptual loss can be written as:
\begin{equation}
\begin{aligned}
L_{per}=L_{1}(\hat{I}_{0},I_0)+L_{MSE}(\hat{I}_{0},I_0).
\end{aligned}
\end{equation}\par
To effectively capture global frequency domain information and reduce recovery bias, we introduce a frequency domain loss that enhances the model's ability to recover the global structure and fine details through joint constraints in both the frequency and spatial domains. Specifically, we apply the fast Fourier transform (FFT) to both the recovered image $\hat{I}_{0}$ and the reference image $I_{0}$ to obtain their frequency domain representations, including magnitude and phase, and compute the $L1$ loss for both the magnitude and phase of the two images. The expression is as follows:
\begin{equation}
\begin{aligned}
A_{\hat{I}_{0}},P_{\hat{I}_{0}} &= \mathcal{F}(\hat{I}_{0})\\
A_{I_{0}},P_{I_{0}} &= \mathcal{F}(I_{0})\\\\
L_{FFT}&=L_1(A_{\hat{I}_{0}},A_{I_{0}})+L_1(P_{\hat{I}_{0}},P_{I_{0}}).
\end{aligned}
\end{equation}
where \(\mathcal{F}\) denotes the Discrete Fourier Transform and where \(A\) and \(P\) represent the amplitude and phase, respectively.
Overall, the final loss function aims to minimize the weighted sum of all these losses:

\begin{equation}
L=\alpha L_{per}+ \lambda  L_{FFT} +  L_{LDG},
\end{equation}
where $\alpha$=2 and $\lambda$=1 are the hyperparameters for each loss term.

\begin{table*}[]\centering
\renewcommand\arraystretch{1.4}

\caption{Quantitative comparison of real and synthetic nighttime flare-corrupted data\cite{10541091}. \textbf{Bold} represents the best result. $^{\dag}$ indicates that this method uses the Flare7k dataset as the training dataset.}
\label{comparsion}
\scalebox{0.55}{
\begin{tabular}{c|c|ccccc|ccccc|c}
\hline
\multirow{2}{*}{Method} & \multirow{2}{*}{Published} &        &       & Real   &        &        &        &       & Synthetic &        &        & \multirow{2}{*}{Param (M)} \\ \cline{3-12}
                        &                            & PSNR   & SSIM  & LPIPS  & G-PSNR & S-PSNR & PSNR   & SSIM  & LPIPS     & G-PSNR & S-PSNR &                            \\ \hline
Sharma et al.$^{\dag}$\cite{sharma2021nighttime}           & CVPR'21                    & 20.492 & 0.826 & 0.1115 & 17.790 & 12.648 & -      & -     & -         & -      & -      & -                          \\
Wu et al.$^{\dag}$\cite{wu2021train}               & ICCV'21                    & 24.613 & 0.871 & 0.0598 & 21.772 & 16.728 & -      & -     & -         & -      & -      & -                          \\
Flare7k$^{\dag}$\cite{dai2022flare7k}        & NeurIPS'22                 & 26.978 & 0.890 & 0.0466 & 23.507 & 21.563 & 27.219 & 0.960 & 0.0241    & 23.981 & 24.365 & 20.45                      \\
Zhou et al.$^{\dag}$\cite{zhou2023improving}             & ICCV'23                    & 25.184 & 0.872 & 0.0548 & 22.112 & 20.543 & 28.779 & 0.939 & 0.0286    & 23.779 & 22.237 & 20.63                      \\
BracketFlare\cite{dai2023nighttime}            & CVPR'23                    & 26.587 & 0.886 & 0.0559 & 23.410 & 22.281 & 28.573 & 0.946 & 0.0297    & 24.573 & 23.682 & 3.64                       \\
IR-SDE\cite{luo2023image}                  & ICML'23                    & 27.121 & 0.891 & 0.0047 & 23.642 & 22.143 & 28.270 & 0.959 & 0.0224    & 23.941 & 22.623 & 34.22                      \\
Retinexformer\cite{cai2023retinexformer}                  & ICCV'23                    &26.142&0.879&0.0630&23.493&19.646& 26.386& 0.935&0.0448&23.103&19.298&1.61                    \\
Flare7k++\cite{10541091}      & TPAMI'24                   & 27.633 & 0.894 & 0.0428 & 23.949 & 22.603 & 29.513 & 0.963 & 0.0209    & 24.724 & 24.188 & 20.45                      \\
Kotp et al.\cite{kotp2024flare}             & ICASSP'24                  & 27.662 & 0.897 & \textbf{0.0422} & 23.987 & 22.847 & 29.573 & 0.961 & 0.0205    & 24.879 & 24.458 & 20.47                      \\
WaveDM\cite{huang2024wavedm}                  & TMM'24                     &    23.892    &  0.827     &   0.0901     &   20.996     &  20.934            &  24.299     &   0.834        &  0.0577      &   21.847     &     20.649     &  156.49                  \\
RDDM\cite{liu2024residual}                    & CVPR'24                    & 24.412 & 0.853 & 0.1560  & 21.422 & 21.285 & 25.256 & 0.873 & 0.1940     & 21.922 & 21.140  & 36.26                      \\
FPro\cite{zhou2024seeing}                   & ECCV'24                     & 26.841 & 0.899 & 0.0516 & 23.936 & 23.405 & 27.928 & 0.948 & 0.0303    & 24.186 & 23.997 & 22.29                     \\ 
CycleRDM\cite{xue2025unified}                    & PR'25                     & 27.198 & 0.887 & 0.0472  & 23.366 & 22.054 & 28.141 & 0.940 & 0.0221    & 24.819 & 23.170 & 22.10                  \\
Reti-Diff\cite{he2025retidiff}                    & ICLR'25                     & 27.097 & 0.893 & 0.0470  & 23.778 & 22.405 & 28.028 & 0.951 & 0.0258    & 24.029 & 22.538 & 26.11                  \\
 \hline
Ours$^{\dag}$                    & -              &  27.428      &   0.895     &  0.0488      &  24.050      &  21.904     &  29.230         &    0.957    & 0.0238   & 24.460 & 21.160   & 23.57                      \\
Ours                    &  -                          & \textbf{28.030} & \textbf{0.903} & \textbf{0.0422} &\textbf{24.509} & \textbf{23.300} & \textbf{30.354} & \textbf{0.967} & \textbf{0.0179}    & \textbf{25.562} & \textbf{25.430} & 23.57                      \\ \hline
\end{tabular}
}

\end{table*}

\begin{table}[]\centering
\caption{Quantitative comparison of real-world nighttime flare-corrupted datasets\cite{10541091}. \textbf{Bold} represents the best result. $^{\dag}$ indicates that this method uses the Flare7k dataset as the training dataset.}
\label{flare-corrupted}
\begin{tabular}{c|c|ccc}
\hline
Methods             & Published & NIQE$\downarrow$ & MUSIQ$\uparrow$ & PI$\downarrow$    \\ \hline
Flare7k$^{\dag}$\cite{dai2022flare7k}             & NeurIPS'22 & 2.725 & 64.157 & 1.856 \\
BracketFlare\cite{dai2023nighttime}          & CVPR'23    & 2.984 & 63.889 & 1.963 \\
IR-SDE\cite{luo2023image}              & ICML'23    & 3.005 & 64.689 & 2.017 \\
Retinexformer\cite{cai2023retinexformer}              & ICCV'23    & 2.815 & 62.350 & 2.097 \\
Flare7k++\cite{10541091} & TPAMI'24   & 2.867 & 64.419 & 1.926 \\
Kotp et al.\cite{kotp2024flare}\               & ICASSP'24  & 2.750  & 64.222 & 1.870  \\ 
RDDM\cite{liu2024residual}                & CVPR'24  & 3.105                    & 60.244                   & 2.110                  \\
FPro\cite{zhou2024seeing}                & ECCV'24  & 2.814                    & 64.901                   & 1.869                  \\
\hline
Ours                & -          & \textbf{2.714} & \textbf{64.702} & \textbf{1.849} \\ \hline
\end{tabular}

\end{table}

\section{Experiments}\label{EXPERIMENTS}
\subsection{Experimental Settings}

\begin{table}[]\centering
\caption{Quantitative comparison of the consumer electronics test dataset\cite{zhou2023improving}. \textbf{Bold} represents the best result. $^{\dag}$ indicates that this method uses the Flare7k dataset as the training dataset.}
\label{consumer}
\begin{tabular}{c|c|lll}
\hline
Methods             & Published  & \multicolumn{1}{c}{NIQE$\downarrow$} & \multicolumn{1}{c}{MUSIQ$\uparrow$} & \multicolumn{1}{c}{PI$\downarrow$} \\ \hline
Flare7k$^{\dag}$\cite{dai2022flare7k}             & NeurIPS'22 & 5.137                    & 58.410                     & 3.579                  \\
BracketFlare\cite{dai2023nighttime}          & CVPR'23    & 4.748                    & 57.643                    & 3.274                  \\
IR-SDE\cite{luo2023image}              & ICML'23    & 5.089                    & 59.213                    & 3.370                   \\
Retinexformer\cite{cai2023retinexformer}              & ICCV'23    & 4.931 & 56.694 & 3.550 \\
Flare7k++\cite{10541091} & TPAMI'24   & 5.029                    & 58.678                    & 3.492                  \\
Kotp et al.\cite{kotp2024flare}                & ICASSP'24  & 4.923                    & 56.913                    & 3.478                  \\
RDDM\cite{liu2024residual}                & CVPR'24  & 5.203                    & 56.874                    & 3.619                 \\
FPro\cite{zhou2024seeing}                & ECCV'24  & 4.860                   & 57.377                   & 3.201                  \\
\hline
Ours                & -          & \textbf{4.646}                    & \textbf{60.137}                    & \textbf{3.085}                  \\ \hline
\end{tabular}

\end{table}
\begin{figure*}[t]
        \centering
        \includegraphics[height=0.88\textwidth, width=\textwidth]{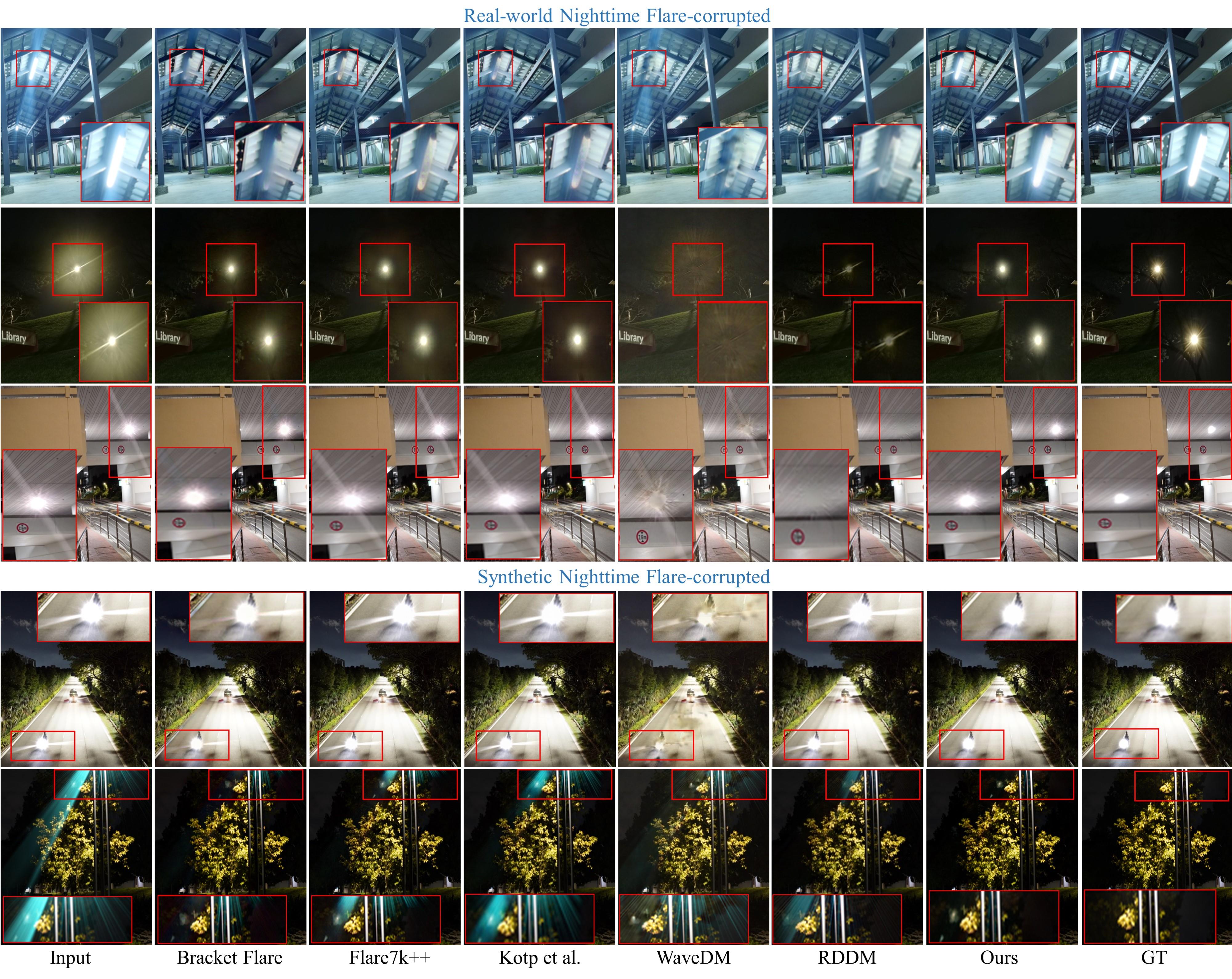}
        \caption{Comparison of the restored results between our method and state-of-the-art methods\cite{dai2023nighttime,10541091,kotp2024flare, liu2024residual, huang2024wavedm} on real and synthetic nighttime flare-corrupted datasets \cite{10541091}. The red box highlights the area with more severe flare artifacts.}
        \label{duihbi1}

    \end{figure*}
\subsubsection{Dataset}
We utilized the Flare7K++\cite{10541091} synthesis pipeline to generate paired flare-corrupted and flare-free images as the training set. Background images were sampled from 24K Flickr images\cite{zhang2018single}. Additionally, flare images and their corresponding light sources were sampled from the Flare7K and Flare-R datasets with a 50$\%$ probability. To ensure a fair comparison, our data augmentation strategy also follows the Flare7K++ protocol. First, inverse gamma correction with $\gamma < U(1.8, 2.2)$ was applied to the flare images (including light sources) and background images to restore linear brightness. For images in our flare dataset, we apply a series of random transformations, including rotation by $U(0,2\pi)$, translation by $U(-300,300)$, shear by $U(-\frac{\pi}{9}, \frac{\pi}{9})$, scaling by $U(0.8,1.5)$, blurring with a kernel size from $U(0.1,3)$, and random flipping, ensuring that paired light source and flare images undergo the same transformations. Additionally, a global color shift sampled from $U(-0.02,0.02)$ is added to the flare image to simulate the effect of flare illuminating the entire scene. For each background image, we apply random RGB scaling with a factor from $U(0.5,1.2)$ and introduce Gaussian noise with variance sampled from a scaled chi-square distribution, $\sigma ^{2}< 0.01\chi ^{2}$. To validate the robustness of our method, we conducted experiments on four test sets: paired test datasets, including the Flare7K++ real and synthetic test datasets\cite{10541091}, and unpaired test datasets, including the consumer electronics test dataset\cite{zhou2023improving} and flare-corrupted dataset\cite{10541091}. The Flare7K++ real and synthetic test datasets both contain 100 flare-corrupted images along with corresponding masks for all the streaks, glare, and light source regions. The flare-corrupted dataset contains 645 images captured by different cameras, some of which are particularly challenging. The consumer electronics test dataset contains damaged images captured by ten different consumer electronics products, and this dataset helps validate the generalizability of the flare removal method.
\subsubsection{Implementation Details} We implemented our framework via PyTorch on a single NVIDIA RTX 3090 GPU. We crop the input images to $512\times512\times3$ and set the batch size to $2$. The Adam optimizer was used with $\beta_1 = 0.9$ and $\beta_2 = 0.99$, and the learning rate was set to $1 \times 10^{-4}$. The learning rate was halved at iteration steps of 150,000 and 200,000.

\subsubsection{Metrics}
In addition to common metrics such as the peak signal-to-noise ratio (PSNR), structural similarity index (SSIM)\cite{wang2004image}, and learned perceptual image patch similarity (LPIPS)\cite{zhang2018unreasonable}, we introduce the S-PSNR and G-PSNR\cite{10541091} to evaluate flare removal performance in the flare and stripe regions. The S-PSNR corresponds to the red area in Fig. \ref{filter}, whereas the G-PSNR corresponds to the yellow area.
These additional metrics provide a more rigorous evaluation of the model's deflare capability. Therefore, we evaluate the performance in our experiments on the basis of the G-PSNR and S-PSNR metrics. To evaluate the performance on the unpaired test datasets, we use three common image quality assessment metrics: NIQE\cite{mittal2012making}, MUSIQ\cite{ke2021musiq} and PI\cite{blau20182018}. These metrics assess image quality from different perspectives. Specifically, NIQE focuses on the naturalness and lack of structural information in the image, MUSIQ evaluates perceptual quality features, and PI reflects the visual quality of the image through perceptual metrics. By combining these evaluations, we can conduct a more comprehensive analysis of the model's performance on the unpaired test set.

\begin{figure*}[t]
        \centering
        \includegraphics[height=0.28\textwidth, width=0.9\textwidth]{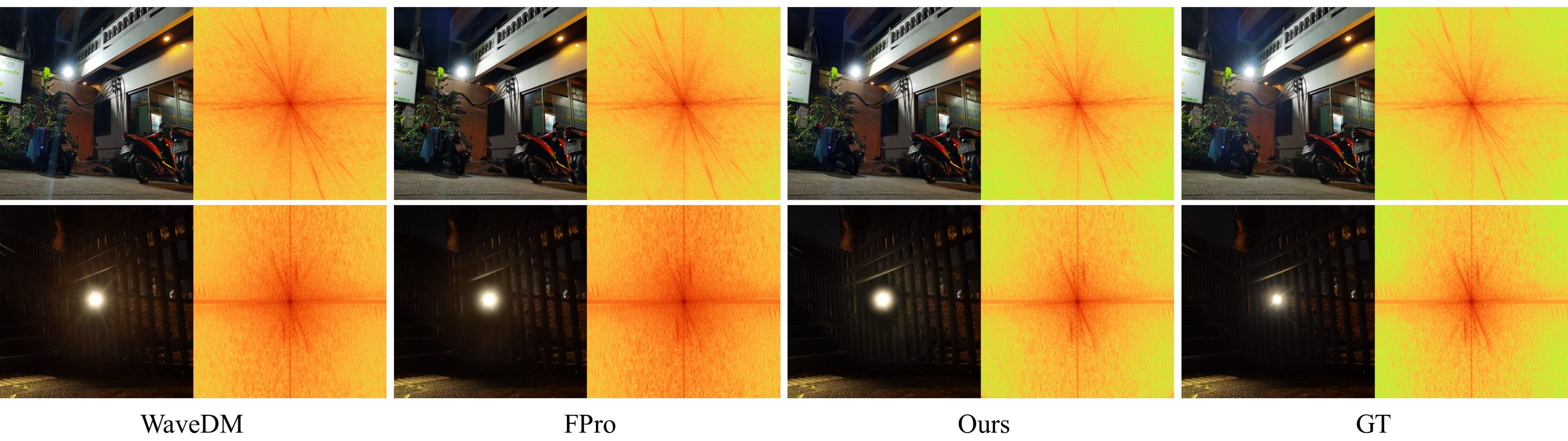}
        \caption{Comparison with existing frequency domain enhanced restoration methods\cite{zhou2024seeing, huang2024wavedm} in both the spatial and frequency domains.}
        \label{freduibi}

\end{figure*}
\begin{figure*}[t]
        \centering
        \includegraphics[height=0.6\textwidth, width=\textwidth]{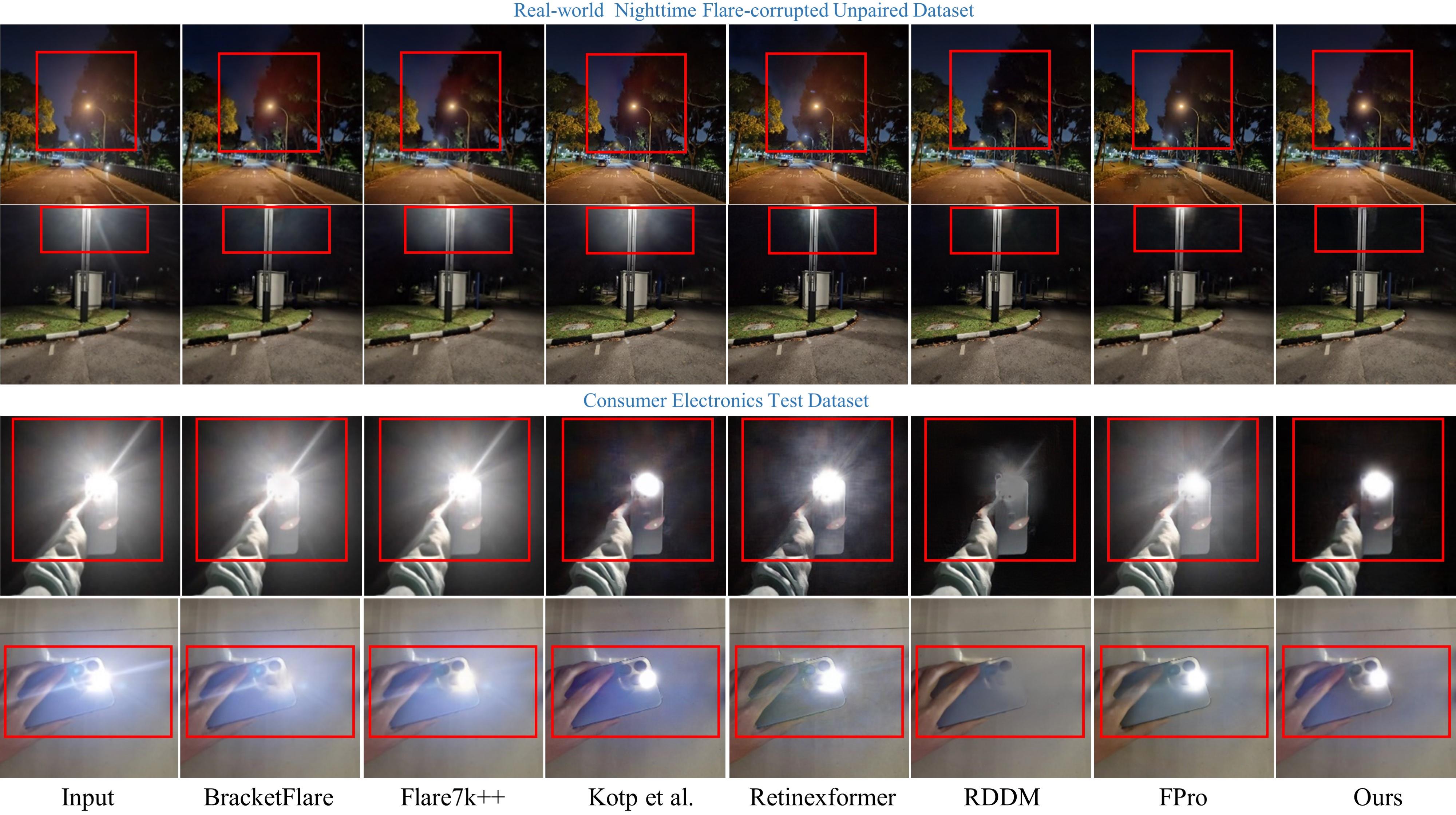}
        \caption{Comparison of the restored results between our method and state-of-the-art methods\cite{dai2023nighttime,10541091} on the real-world nighttime flare-corrupted dataset\cite{10541091} and consumer electronics test dataset\cite{zhou2023improving}. The red box highlights the area with more severe flare artifacts.}
        \label{duihbi}

    \end{figure*}

\subsection{Experimental Results}
\subsubsection{Quantitative Comparison}
\textbf{Paired Test Datasets:} Table. \ref{comparsion} presents the comparison results between our method and other state-of-the-art methods on the paired datasets and the Flare7k real and synthetic test datasets. The results clearly show that our method achieves optimal performance across all the metrics. Specifically, on both paired datasets, our method demonstrates significant improvements in the general metrics PSNR and SSIM compared with the current state-of-the-art methods\cite{10541091,kotp2024flare, luo2023image}, proving its effectiveness in improving the quality of restored images. Our method also shows noticeable enhancements for the evaluation metrics targeting flare regions, G-PSNR, and S-PSNR proposed by Flare7k++\cite{10541091}. On the synthetic test dataset, our G-PSNR and S-PSNR increased by 0.838 dB and 1.242 dB, respectively, from 24.724 dB and 24.188 dB, respectively. For the real test set, our G-PSNR and S-PSNR improved by 0.56 dB and 0.697 dB, respectively, from 23.949 dB and 22.603 dB, respectively. Compared with other frequency domain enhanced image restoration methods, such as WaveDM\cite{huang2024wavedm} and FPro\cite{zhou2024seeing}, our method achieves significant advantages across all evaluation metrics. These results indicate that our method demonstrates outstanding performance in flare removal.\par
\textbf{Unpaired Test Datasets:} First, Table \ref{comparsion} presents the comparison results of our method with other state-of-the-art methods\cite{dai2022flare7k,10541091,kotp2024flare, luo2023image} on the challenging dataset proposed by \cite{10541091}. Our method achieves optimal performance across various perceptual metrics, demonstrating that our model performs robustly even when removing images severely degraded by flares. To further demonstrate the robustness of our method, we conducted experiments on the consumer electronics test dataset proposed by \cite{zhou2023improving}, with the results shown in Table \ref{consumer}. The results indicate that our method also performs well in removing flares captured by smartphone cameras.
\subsubsection{Qualitative
Comparisons}
\textbf{Paired Test Datasets:} We first present a visual comparison of the Flare7k++ real and synthetic test datasets in Fig. \ref{duihbi1}. The results demonstrate that our method is the most effective at eliminating large-scale artifacts and restoring local details around light sources. As shown in the first row of Fig. \ref{duihbi1}, existing methods often damage the local details of the light source after removing the flare, whereas our method successfully restores the content in these regions. In the second row of Fig. \ref{duihbi1}, existing methods remove glare effects but leave behind significant flare artifacts, whereas our method achieves satisfactory results. As shown in Fig. \ref{freduibi}, our method demonstrates superior performance in nighttime flare removal tasks compared with existing frequency-domain enhancement methods such as WaveDM\cite{huang2024wavedm} and FPro\cite{zhou2024seeing} in both the spatial and frequency domains. In the frequency domain, our method not only preserves the mid- and high-frequency structural features of the image but also effectively suppresses low-frequency noise introduced by flares. The frequency spectrum distribution is more consistent with the GT, with a well-balanced central intensity and clearly defined texture distribution, avoiding the energy distribution anomalies or blurring problems observed in other methods. Overall, the superior performance of our method in both the spatial and frequency domains verifies its advantages in structural restoration, spectral consistency, and detail preservation, fully demonstrating its robustness in complex nighttime lighting scenarios.\par
\textbf{Unpaired Test Datasets:} Next, we present a visual comparison of the unpaired test dataset, flare-corrupted images\cite{10541091}, as shown in Fig. \ref{duihbi}. Our method also delivers excellent restoration performance for real-world nighttime images with flare corruption. To further verify the robustness of our approach, we conducted tests on the consumer electronics test dataset\cite{zhou2023improving}, as shown in Fig. \ref{duihbi}. The results demonstrate that our method exhibits strong generalization ability, effectively removing daytime flares and intense flares caused by smartphone lenses, particularly the severe flares generated by smartphone lenses, as shown in the last two columns of Fig. \ref{duihbi}.

% \begin{figure*}[t]
%         \centering
%         \includegraphics[height=0.5\textwidth, width=\textwidth]{duibi2.pdf}
%         \caption{Visual comparison of the results from state-of-the-art methods\cite{dai2023nighttime,10541091} on the real-world nighttime flare-corrupted dataset\cite{10541091}. The red box highlights the area with more severe flare artifacts.}
%         \label{duihbi2}
%     \vspace{-1em}
%     \end{figure*}
% \begin{figure*}[t]
%         \centering
%         \includegraphics[height=0.5\textwidth, width=\textwidth]{duibi3.pdf}
%         \caption{Visual comparison of the results from state-of-the-art methods\cite{dai2023nighttime,10541091} on the consumer electronics test dataset\cite{zhou2023improving}. The red box highlights the area with more severe flare artifacts.}
%         \label{duihbi3}
%     \vspace{-1em}
%     \end{figure*}

\begin{table*}[]\centering
\caption{Ablation study on the combination of different modules in our network. \textbf{Bold} represents the best result. '$\checkmark$' means to use the module, and ‘$\times$’ means not use it.}
\label{ablation}
\renewcommand\arraystretch{1.2}
\scalebox{0.65}{
\begin{tabular}{cccccccccl}
\cline{1-9}
\multicolumn{3}{c}{Training sets} & \multicolumn{6}{c}{Flare7k++ real}                             &  \\ \cline{1-9}
GDFG   & LDGM   & Frequency Loss  & PSNR$\uparrow$   & SSIM$\uparrow$  & LPIPS$\downarrow$  & G-PSNR$\uparrow$ & S-PSNR$\uparrow$ & Inference Time (s) &  \\ \cline{1-9}
$\times$      & $\times$      & $\times$               & 27.633 & 0.894 & 0.0428 & 23.949 & 22.603 & 0.285              &  \\
$\checkmark$      & $\times$      & $\times$               & 27.740  & 0.899 & 0.0425 & 24.310  & 23.159 & 0.298              &  \\
$\times$      & $\checkmark$      & $\times$               & 27.652 & 0.898 & 0.0458 & 24.179 & 23.121 & 0.286              &  \\
$\checkmark$      & $\times$      & $\checkmark$               & 27.951 & 0.901 & 0.0429 & 24.492 & 23.261 & 0.298              &  \\
$\checkmark$      & $\checkmark$      & $\checkmark$               & \textbf{28.030}  & \textbf{0.903} & \textbf{0.0422} & \textbf{24.509} & \textbf{23.300}   & 0.303              &  \\ \cline{1-9}
\end{tabular}
}
\end{table*}
\begin{table}[]\centering

\caption{Ablation study of different filter dimensions N in the GDFG. \textbf{Bold} represents the best result.}
\label{Nab}
\scalebox{0.9}{
\begin{tabular}{cccccc}
\hline
Dimensions $N$ & PSNR$\uparrow$   & SSIM$\uparrow$   & LPIPS$\downarrow$  & G-PSNR$\uparrow$ & S-PSNR$\uparrow$ \\ \hline
2 & 27.891 & 0.899  & 0.0426 & 24.257 & 23.138 \\
3 & 27.958 & 0.901  & 0.0426 & 24.435 & \textbf{23.355} \\
4 & \textbf{28.030}  & \textbf{0.903} & \textbf{0.0422} & \textbf{24.509} & 23.300   \\
5 & 27.991 & 0.901  & 0.0423 & 24.415 & 23.275 \\ \hline
\end{tabular}
}

\end{table}
\subsection{Ablation Study}

\begin{figure}[t]
        \centering
        \includegraphics[height=0.42\textwidth, width=0.8\textwidth]{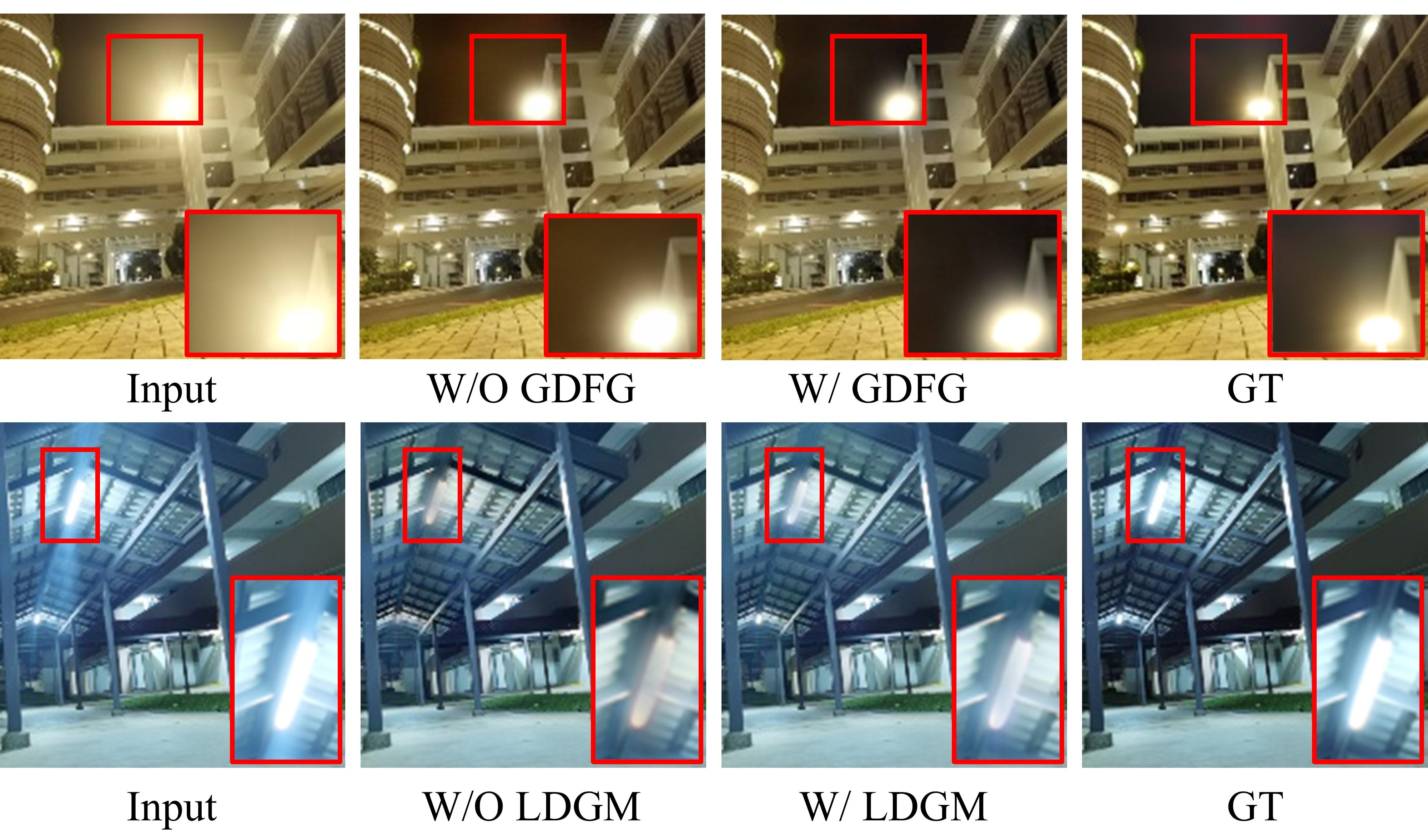}
        \caption{Comparison of image restoration results with and without the Global Dynamic Frequency-domain Guidance Module and the Local Detail Guidance Module.}
        \label{visab}

    \end{figure}

\begin{figure}[t]
        \centering
        \includegraphics[height=0.5\textwidth, width=0.5\textwidth]{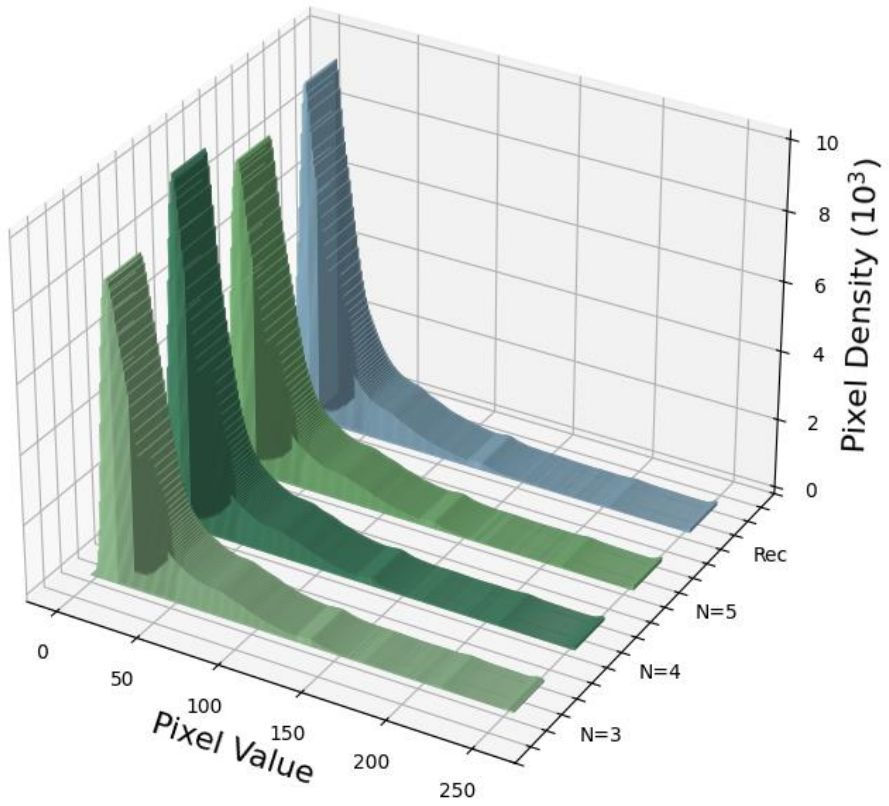}
        \caption{Comparison of the pixel value distributions between the restored image and the reference image for different filter dimensions $N$.}
        \label{N}

    \end{figure}

\subsubsection{Influence of the Global Dynamic Frequency-domain Guidance Module}
Nighttime flares are often accompanied by large-scale artifacts, which are challenging to eliminate by existing methods. To address this issue, we design the GDFG module to dynamically separate flare artifacts. As shown in Fig. \ref{duihbi1}, our method effectively mitigates this problem. To further verify the contribution of this module to our network, we conducted ablation experiments, as presented in Table \ref{ablation}. The results demonstrate that the GDFG module significantly enhances the model's flare removal performance. As shown in Fig. \ref{visab}, we compare the restoration results with and without the global dynamic frequency-domain guidance module. The introduction of the GDFG module significantly suppresses large-scale artifacts around the light source and enhances the clarity of local details in its vicinity. These results demonstrate that the GDFG module effectively mitigates flare interference and improves local detail restoration, thereby substantially enhancing the overall image quality. If the frequency domain loss we designed is added to the end of the network, the perceptual quality of the image will be further improved through the joint constraints of the spatial and frequency domains.\par
\begin{figure}[t]
        \centering
        \includegraphics[height=0.38\textwidth, width=0.5\textwidth]{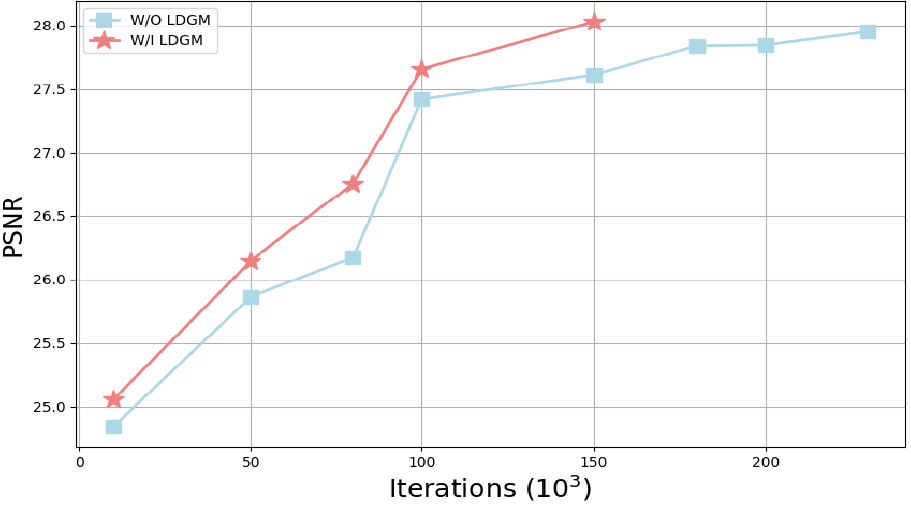}
        \caption{Visualization of model convergence speed with and without the LDGM.}
        \label{speed}

    \end{figure}
We further visualized the ability of the GDFG module to perceive flares, as shown in Fig. \ref{filter}.
The figure illustrates Flare7k++, our method with and without the GDFG module, as well as annotations of the flare regions. The results indicate that the GDFG module effectively perceives flare and artifact regions, highlighting its significant contribution to flare detection and removal. Furthermore, we conducted an ablation study on the filter dimensions within the module, with the results presented in Table \ref{Nab}. When the dimension is set to $N=4$, the model performs best. Therefore, $N=4$ is selected as the default configuration in this study. Moreover, as shown in Fig. \ref{N}, we compared the pixel distributions of the restored images under different filter dimensions with those of the reference image. When $N=4$, the pixel distribution of the restored image aligns most closely with the reference image, clearly demonstrating that the model achieves optimal perceptual capability under this configuration.

% \begin{table}[]
% \caption{Ablation studies of the loss function terms.}
% \centering
% % \renewcommand\arraystretch{1.3}
% \scalebox{1.1}{
% \begin{tabular}{l|cc}
% \hline
%              & PSNR↑            & SSIM↑                                       \\ \hline
% without $\mathcal{L}_{c}$ & 28.131                                 & 0.895                                             \\
% without  $\mathcal{L}_{p}$    & 25.478                                       &0.853                        \\
% without $\mathcal{L}_{s} $     & 28.979                                 & 0.902                                \\
% without   $\mathcal{L}_{m}$   & 28.305                                       & 0.900                          \\ \hline
% Default      & { \textbf{29.857}} & {\textbf{0.908}} \\ \hline
% \end{tabular}
% }
% \label{loss}
% \end{table}
% Table 2 demonstrates the effectiveness of the designed text-driven appearance reconstruction method. The performance of the model is effectively improved by connecting the Fourier frequency domain through multimodal semantics and dynamically adjusting the weights of perceptual loss.

\begin{table}[]\centering
\caption{Ablation of the temperature hyperparameter $\tau$ in the local detail guidance module. \textbf{Bold} represents the best result.}
\label{ab_t}
\begin{tabular}{c|cccc}
\hline
Metrics & $\tau$=0.05 & $\tau$=0.07 & $\tau$=0.10  & $\tau$=0.20  \\ \hline
G-PSNR  & 24.333 & \textbf{24.509} & 24.010 & 23.917 \\
S-PSNR  & 23.189 & \textbf{23.300} & 22.939 & 22.845 \\ \hline
\end{tabular}
\end{table}

\begin{table*}[]\centering
\caption{Ablation study on the perceptual loss weight coefficient $\alpha$ and the frequency loss weight coefficient $\lambda$. \textbf{Bold} represents the best result.}
\label{ab_loss}

\scalebox{0.8}{
\begin{tabular}{c|cccc|cccc}
\hline
Metrics & $\alpha$=1    & $\alpha$=2    & $\alpha$=3    & $\alpha$=5    & $\lambda$=0.5  & $\lambda$=1.0  & $\lambda$=1.5  & $\lambda$=2.0  \\ \hline
G-PSNR  & 24.149 & \textbf{24.509} & 24.338 & 24.090 & 24.477 & \textbf{24.509} & 23.990 & 24.325 \\
S-PSNR  & 23.075 & \textbf{23.300} & 23.226 & 23.261 & 22.864 & \textbf{23.300} & 23.053 & 23.110 \\ \hline
\end{tabular}
}
\end{table*}

\subsubsection{Influence of the local detail guidance module}
To address the issue of content loss caused by flare removal, we design the LDGM module on the basis of a contrastive learning strategy, aiming to maximize the mutual information between the localized light source regions of the restored image and the reference image. We conducted ablation experiments on this module, as shown in Table \ref{ablation}. The results indicate that introducing this module significantly improves the S-PSNR value, with an increase of 0.607 compared with the baseline method. This demonstrates the module's effectiveness in restoring flare streak regions. Additionally, the G-PSNR metric also shows an improvement, verifying the module's positive impact on recovering content near the light source.\par

We present visual comparisons of the restored results with and without this module in Fig. \ref{visab}. The visualization clearly shows that some details in the light source region are lost due to flare removal, whereas with the LDGM module, our method effectively restores these details. Furthermore, we evaluated the convergence speed of the model, as illustrated in Fig. \ref{speed}. With the inclusion of the LDGM module, our model achieves optimal performance at 150,000 iterations, which is nearly 40$\%$ faster than without the module.\par
\begin{figure}[t]
        \centering
        \includegraphics[height=0.25\textwidth, width=0.8\textwidth]{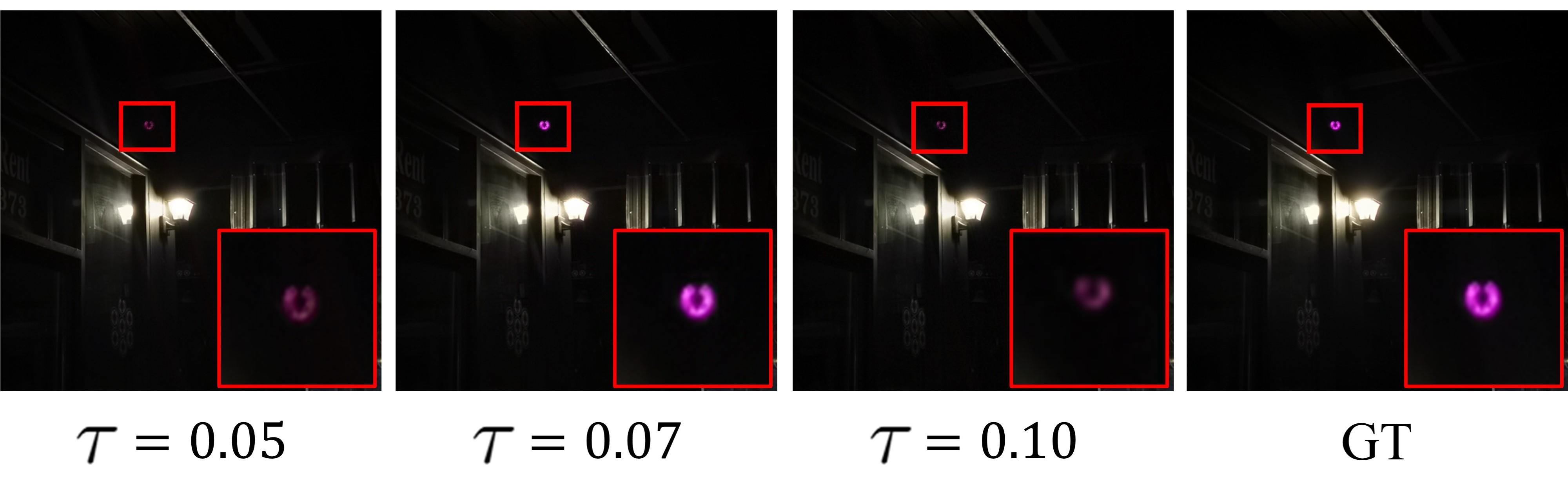}
        \caption{Comparison of the restoration results under different temperature hyperparameters $\tau$.}
        \label{visab_t}

    \end{figure}

To investigate the impact of the temperature parameter $\tau$ in the local detail guidance module (LDGM), we conducted an ablation study, as shown in Table \ref{ab_t}. The experimental results indicate that the model achieves the best performance when $\tau$=0.07. This is because the temperature parameter controls the smoothness of the Softmax distribution, thereby affecting the discrimination strength between positive and negative sample pairs. When $\tau$ is too small (e.g., 0.05), the similarity score differences are excessively amplified, making the loss function overly sensitive to minor perturbations and increasing training instability. Conversely, when $\tau$ is too large (e.g., 0.1 or 0.2), the similarity distribution becomes overly smooth, weakening the distinction between positive and negative sample pairs and thus reducing the discriminative power of the features. As shown in Fig. \ref{visab_t}, we compare the restoration results under different temperature parameter $\tau$ settings. When $\tau$=0.07, the restored result is closest to the ground truth. Overall, setting $\tau$=0.07 strikes a balance by sufficiently emphasizing positive sample pairs and suppressing negative ones while maintaining training stability and effectively improving model performance.
\subsubsection{Influence of GDFG and LDGM modules on global and local flare removal}
\begin{figure}[t]
        \centering
        \includegraphics[height=0.25\textwidth, width=0.8\textwidth]{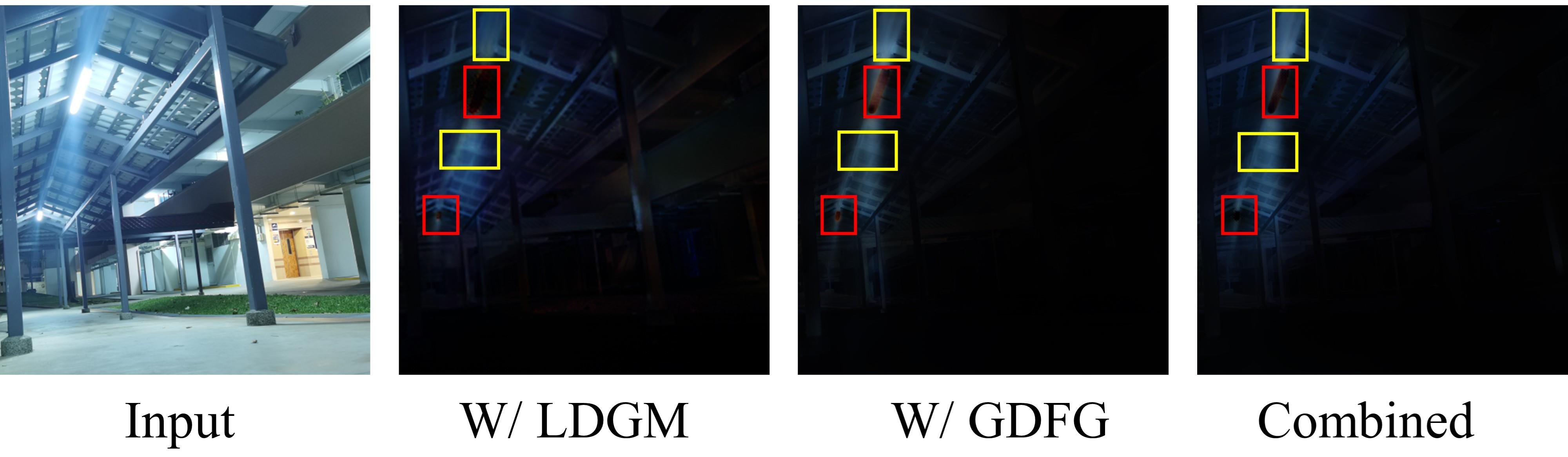}
        \caption{ Comparison of predicted flares using different configurations of DFDNet. Yellow boxes highlight regions where GDFG enhances the model’s ability to identify flare artifacts. However, without LDGM, the model tends to misclassify light sources as flare regions (red boxes). LDGM alone avoids light source misclassification but struggles to fully detect flare artifacts. The combination of both modules yields the most accurate prediction by balancing global guidance and local structural refinement.}
        \label{g_l}

    \end{figure}

To validate the complementary benefits of our global and local guidance modules, we visualize the predicted flare maps under three different settings: using only LDGM, only GDFG, and the combination of both. As shown in Fig. \ref{g_l}, when only the GDFG module is used, the network captures the global frequency-based flare patterns more completely (yellow boxes), but often misclassifies the light source as a flare region (red boxes). In contrast, LDGM effectively suppresses the misidentification of light sources but struggles to accurately delineate complete flare structures. When both modules are integrated, the network not only correctly identifies flare artifacts but also preserves light sources, leading to the most accurate prediction. This demonstrates the strong synergy between the frequency-domain global guidance and the spatial-domain local contrastive learning in our proposed DFDNet.

\subsubsection{Influence of loss weight coefficient}
We conducted an ablation study on the weighting coefficients of the perceptual loss $\alpha$ and the frequency loss $\lambda$, as shown in Table. \ref{ab_loss}. The network achieves the best performance when $\alpha$=2 and $\lambda$=1.0. Therefore, we set $\alpha$=2 and $\lambda$=1.0 as the default configurations in our experiments.
\begin{figure}[t]
        \centering
        \includegraphics[height=0.45\textwidth, width=0.78\textwidth]{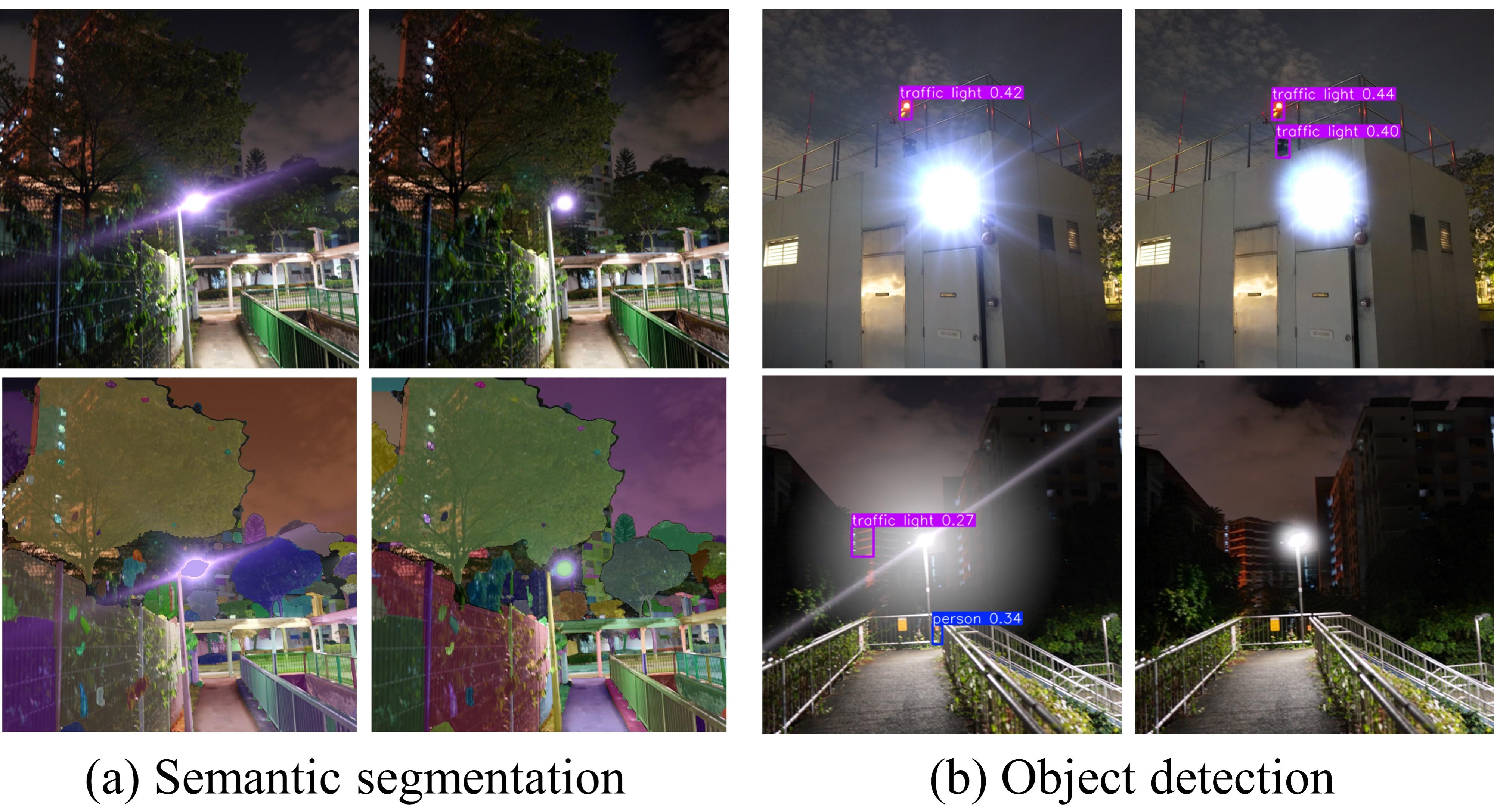}
        \caption{Application of the proposed flare removal algorithm to downstream tasks.}
        \label{downtask}

    \end{figure}
\subsection{Flare Removal for Downstream Tasks}
As illustrated in Fig. \ref{downtask}, our method clearly demonstrates advantages in downstream applications. We first perform object detection on degraded and restored images via YOLOv11 (see Fig. \ref{downtask}(a)). It can be observed that flare removal significantly improves nighttime detection performance. Specifically, in degraded images, some objects occluded by flares are completely missed by the detector. Moreover, the presence of intense flares leads the detector to mistakenly identify certain background regions as foreground objects due to the distortion of visual cues. In contrast, our flare removal approach effectively restores object details and suppresses misleading artifacts, resulting in more accurate and reliable detection outcomes. Additionally, Fig. \ref{downtask}(b) presents the results of a semantic segmentation task using SAM\cite{kirillov2023segment}, further validating the effectiveness of our method in enhancing the performance of downstream applications.
\begin{figure}[t]
        \centering
        \includegraphics[height=0.3\textwidth, width=0.7\textwidth]{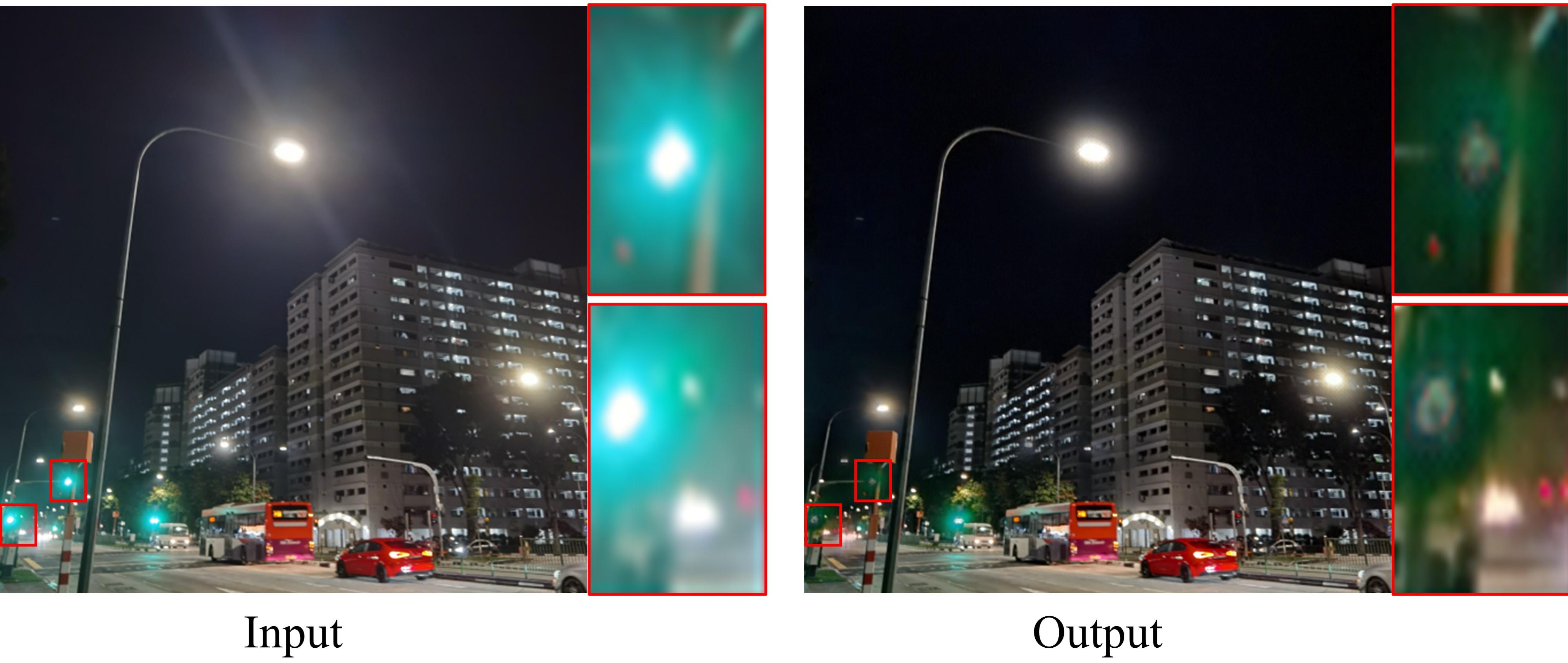}
        \caption{Limitations of our method.}
        \label{limitions}

    \end{figure}

\section{Limitations and Future Work}\label{Limitations}
The experimental results demonstrate that DFDNet exhibits outstanding performance in the task of flare removal. However, this study has several limitations. Specifically, some small light sources may suffer from detail loss, as shown in Fig. \ref{limitions}. This issue is also present in existing methods and remains unsolved. The possible reason lies in the process of removing large-area flares, where these small light sources are mistakenly identified as reflected flares and removed together, resulting in the loss of fine details. To address this limitation in future work, we plan to introduce prior information to enhance the preservation of key texture features. Additionally, we explore the integration of our method with more advanced deep priors, such as the Segment Anything Model\cite{kirillov2023segment}, Depth Anything Model\cite{yang2024depth}. We also intend to extend the application of our method to a wider range of image processing tasks, including image raindrop removal, image shadow removal, and image inpainting, to further improve its practicality and generalizability capabilities.\par
Furthermore, deploying flare removal methods on mobile devices or other resource-constrained platforms presents significant technical challenges, such as excessive model size or high computational resource requirements, which limit their wide application in practical scenarios to some extent. To address these issues, future research should focus on exploring more efficient network structures and optimization strategies while also aiming to design a lightweight flare removal network to further increase its practicality and deployability.

\section{Conclusion}\label{Conclusions}
In this paper, we propose the dynamic frequency-guided deflare network (DFDNet) to address the challenges of flare artifacts in nighttime photography. DFDNet learns flare characteristics from a frequency domain perspective and integrates a contrastive learning strategy to align local features, effectively removing large-scale artifacts and significantly reducing the structural damage caused by strong light sources in local areas. The proposed global dynamic frequency-domain guidance (GDFG) module uses frequency-domain information to dynamically optimize global features, enabling the precise decoupling of flare information from content information and minimizing interference with surrounding content. Additionally, we design a local detail guidance module (LDGM) based on a contrastive learning strategy to align local features with the reference image, reduce the loss of local details, and ensure fine-grained image restoration. The experimental results demonstrate that DFDNet performs superior flare removal and image quality restoration, making outstanding contributions to flare removal tasks.

\bibliography{PR}
\bibliographystyle{plain}
%% For numbered reference style

%   \textit{\LaTeX: a document preparation system},
%   Addison Wesley, Massachusetts,

% \end{thebibliography}
\end{document}